\documentclass[conference]{IEEEtran}
\usepackage[utf8]{inputenc}
\usepackage[T1]{fontenc}
\usepackage{amsmath, amssymb, amsfonts}
\usepackage{algorithmic}
\usepackage{graphicx}
\usepackage{cite}
\usepackage{textcomp}
\usepackage{nicefrac}
\usepackage{microtype}
\usepackage{booktabs}
\usepackage[dvipsnames]{xcolor}
\usepackage[colorlinks=True, linkcolor=black, urlcolor=blue, citecolor=blue]{hyperref}
\usepackage{tikz}

\usepackage{comment}

\def\BibTeX{{\rm B\kern-.05em{\sc i\kern-.025em b}\kern-.08em
    T\kern-.1667em\lower.7ex\hbox{E}\kern-.125emX}}

\begin{document}

\title{On the Unreasonable Efficiency of State Space Clustering in Personalization Tasks}

\author{
    \IEEEauthorblockN{Anton Dereventsov}
    \IEEEauthorblockA{\textit{Lirio AI Research}\\
    \textit{Lirio, LLC}\\
    Knoxville, TN, USA\\
    adereventsov@lirio.com}
\and
    \IEEEauthorblockN{Raju Vatsavai}
    \IEEEauthorblockA{\textit{Behavioral Reinforcement Learning Lab}\\
    \textit{Lirio, LLC}\\
    Knoxville, TN, USA\\
    rvatsavai@lirio.com}
\and
    \IEEEauthorblockN{Clayton G.~Webster}
    \IEEEauthorblockA{\textit{Behavioral Reinforcement Learning Lab}\\
    \textit{Lirio, LLC}\\
    Knoxville, TN, USA\\
    cwebster@lirio.com}
}

\maketitle
%%%%%%%%%%%%%%%%%%%%%%%%%%%%%%%%%%%%%%%%%%%%%%%%%%%%%%%%%%%%%%%%%%%%%%%%%%%%%%%%%%%%%%%%%%%%%%%%%%%%%%%

%%%%%%%%%%%%%%%%%%%%%%%%%%%%%%%%%%%%%%%%%%%%%%%%%%%%%%%%%%%%%%%%%%%%%%%%%%%%%%%%%%%%%%%%%%%%%%%%%%%%%%%
\begin{abstract}
In this effort we consider a reinforcement learning (RL) technique for solving personalization tasks with complex reward signals.
In particular, our approach is based on state space clustering with the use of a simplistic $k$-means algorithm as well as conventional choices of the network architectures and optimization algorithms.  Numerical examples demonstrate the efficiency of different RL procedures and are used to illustrate that this technique accelerates the agent's ability to learn and does not restrict the agent's performance.
\end{abstract}
%%%%%%%%%%%%%%%%%%%%%%%%%%%%%%%%%%%%%%%%%%%%%%%%%%%%%%%%%%%%%%%%%%%%%%%%%%%%%%%%%%%%%%%%%%%%%%%%%%%%%%%

%%%%%%%%%%%%%%%%%%%%%%%%%%%%%%%%%%%%%%%%%%%%%%%%%%%%%%%%%%%%%%%%%%%%%%%%%%%%%%%%%%%%%%%%%%%%%%%%%%%%%%%
\section{Introduction}\label{sec:intro}
Conventional personalization focuses on personal, transactional, demographic, and possibly health-related information, such as an individual's name, age, residential location, employment, purchases, medical history, etc.
The most basic example of this task involves the inclusion of the customer's name in the subject line and/or the content of an email.
This technique relies on generalization and profiling to make specific assumptions about the individual based on their characteristics.
Additional applications of personalization include: web content personalization and layout customization~\cite{10.1016/j.jss.2016.02.008, Ricci2011IntroductionTR}; customer-centric interaction with their healthcare providers~\cite{lasalvia2020personalization, 10.1145/3318236.3318249}; personalized medical treatments~\cite{2007_aspinall, 10.1016/S0167-7799(01)01814-5}; and music, video, and movie recommender systems~\cite{10.1145/2623372}.

Hyper-personalization, on the other hand, is a more complex and transformative approach that considers behavioral and real-time data produced as the result of user action.
Examples of this technique include online browsing behavior, communication history, in-application choices, and other engagement data, which is used to interpret the user's intent.
This allows for more personalized experiences in the form of, for instance, adding individualized engaging sections to the body of an email or push notifications at the time when the customer is typically active, which result in more customized communication and thus, ultimately, greater conversion.

Reinforcement learning (RL) approaches have been increasingly applied to personalization tasks~\cite{10.1109/TNNLS.2020.2975035, 10.1038/s41591-018-0310-5, 10.1007/s11257-021-09292-w, Zhao2009ReinforcementLD, pmlr-v85-yauney18a, Yu2019ReinforcementLI, CORONATO2020101964}.
The combination with clustering techniques has also been studied in the works~\cite{10.1007/978-3-540-95995-3_14, 10.1038/s41539-019-0054-0, 10.1145/2566486.2567991}.
Implementation of the clustering methods typically simplifies the problem and allows an RL agent to be trained with fewer data points, which is essential in real-world applications.
Nonetheless, utilizing any form of clustering unavoidably leads to a loss of information about the environment, which might prevent an agent from solving the task completely.
This trade-off between the environment's complexity and representation gives rise to the immediate question of which RL algorithms and clustering methods lead to a satisfactory performance.

In this paper we explore the unexpected efficiency of a naive approach to state space clustering on complex personalization tasks.
For a clearer demonstration of this phenomenon, we consider the most simplistic $k$-means approach (see e.g.~\cite{Bock2007, 10.5013/IJSSST.a.17.24.06} and the references therein) for clustering that does not account for the reward distribution and is based solely on the state representation as a vector in the state space.
Our main contributions are:
\begin{itemize}
    \item definition and formalization of the synthetic personalization setting;
    \item development of the framework for constructing a wide range of synthetic personalization environments; and
    \item empirical comparison between the clustering-based and full RL-approaches for solving the personalization tasks.
\end{itemize}
%%%%%%%%%%%%%%%%%%%%%%%%%%%%%%%%%%%%%%%%%%%%%%%%%%%%%%%%%%%%%%%%%%%%%%%%%%%%%%%%%%%%%%%%%%%%%%%%

%%%%%%%%%%%%%%%%%%%%%%%%%%%%%%%%%%%%%%%%%%%%%%%%%%%%%%%%%%%%%%%%%%%%%%%%%%%%%%%%%%%%%%%%%%%%%%%%
\section{Background}\label{sec:background}
In the current work we address a personalization task with a reinforcement learning approach.
Specifically, we view the personalization task as a contextual bandit problem, introduced in~\cite{langford2007epoch}. For convenience of presentation we use the notational standard MDPNv1~\cite{thomas2015notation}. Namely, $\mathcal{S}$ denotes the state (context) space, $\mathcal{A}$ denotes the action space, $\mathcal{R} \subset \mathbb{R}$ denotes the reward space, and $r : \mathcal{S} \times \mathcal{A} \to \mathcal{R}$ denotes the reward function. We assume that the reward space $\mathcal{R}$ is bounded and that the reward function $r$ is deterministic.

Though reinforcement learning has been found to be highly successful in games, control, and personalization applications (see e.g.~\cite{vinyals2019grandmaster,berner2019dota}), it suffers from several limitations. In particular, many RL algorithms require samples that are polynomial in the size of the state space~\cite{Strehl-09}. Although the polynomial sample complexity may not be a challenge anymore with advances in computing capabilities, the size of a state space has been a problem for an MDP, as it grows super-polynomially with the number of dimensions (or attributes)~\cite{Abel-16}. This curse of dimensionality has been addressed in the literature through state space abstraction.

In~\cite{Giunchiglia-92}, abstraction is defined as a mapping from one problem representation to a new representation, while preserving some properties of the original representation. Thus, learning a state abstraction involves finding a mapping from the original state space $\mathcal{S}$ to another more compact space $\bar{\mathcal{S}}$. Abstraction has been widely studied in the literature to both overcome the curse of dimensionality as well as to exploit the computational advantages it offers for RL algorithms. Abstraction is applicable to both temporal and state space. In this work, we focus mostly on state space abstraction. 

Several techniques have been proposed in the literature to find a compact state space $\bar{\mathcal{S}}$, they can be broadly grouped into: (i) feature selection, (ii) unsupervised learning (clustering), and (iii) hierarchical approaches. 

Feature selection methods can start from a null set and gradually add relevant state attributes, or start from a complete set of state attributes and gradually remove unnecessary attributes. These methods are computationally expensive and also require huge amounts of data. A more efficient feature selection  method is proposed in~\cite{Konidaris-09}, where the algorithm starts with a fixed number of potential state space abstractions and incrementally selects an abstraction based on an approach similar to model selection using Bayesian Information Criterion (BIC). Several other approaches have been proposed to selectively construct compact and accurate feature spaces~\cite{Kolter-09,Thanh-13,Wookey-15} in recent years. 

On the other hand, clustering based approaches seek to construct a new compact state space that preserves important properties such as topology or the structure of the underlying physical system. When performing clustering, one has to pay attention to the novelty (typically measured in terms of counts, empowerment, agents belief, or prediction error) and quality in the neighboring regions of the current state. This observation is the principle behind the Clustered Reinforcement Learning (CRL) algorithm proposed in~\cite{Ma-19} to facilitate  better exploration by the agent. CRL uses a K-means clustering to divide the collected states into several clusters followed by a novel bonus reward that considers both novelty and quality in the neighboring area, i.e. states which share the same cluster with the current state. Experiments on continuous control tasks and Atari games showed that CRL performed better than the trust region based policy gradient method and locality sensitive hashing based methods~\cite{Schulman-15,Tang-16}. In the recent work~\cite{Mandel-16}, the authors argue that speeding up an agent's learning requires generalization of experience across states and propose a Thompson Clustering for Reinforcement Learning (TCRL) framework for reinforcement learning in discrete MDPs with a medium/small state space. TCRL showed better performance on various domains, including the cases where no states are similar. 

In recent years, a great deal of hierarchical reinforcement learning (HRL) algorithms (see~\cite{Dietterich-00,Ghavamzadeh-01,HernandezGardiol-00,Mannor-04,Alexey-19,Guo-16,Rafati-19}) were developed to address the challenges posed by huge state spaces and sparse delayed reward feedback. The hierarchical approaches decompose a complex task into a set of sub-tasks (or subgoals) using hierarchical structures. Once useful subgoals are discovered, an HRL agent should be able to learn the skills to attain those subgoals. However, finding a good decomposition and subgoals is still a major challenge.  In~\cite{Alexey-19}, authors propose an algorithm for automatically building a hierarchy by dividing the state space into clusters and allocating sub-targets by the modified bottleneck method. The states recommended for visiting are called bottlenecks. Some of these challenges were addressed in a model-free HRL~\cite{Rafati-19}. In this model-free HRL framework, subgoals were discovered using unsupervised learning and anomaly detection methods. More detailed approaches can be found in a recent survey paper~\cite{Pateria-21}.

A particular instance of state abstraction in the form of sophisticated state space aggregations is considered in~\cite{Li-06,Abel-16}, where the authors explore the question of learning from such abstractions. In~\cite{Abel-16}, authors investigate several approximate state abstractions and present theoretical guarantees of the quality of behaviors derived from these approximate abstractions. Experimental results show that the approximate abstractions lead to a reduction in task complexity. In~\cite{Li-06} the authors study five different abstraction schemes and provide a unified treatment of state abstraction for Markov decision processes. Unlike the clustering considered in this work, these abstractions treat the state space not in isolation but as a part of the environment and in certain cases can be used to learn the optimal policy. On the other hand, our work is focused on building a general purpose simulation and experimentation framework that allows for the empirical comparison of various abstraction strategies using clustering-based methods as well as the exploration of general RL algorithms for solving personalization tasks.
%%%%%%%%%%%%%%%%%%%%%%%%%%%%%%%%%%%%%%%%%%%%%%%%%%%%%%%%%%%%%%%%%%%%%%%%%%%%%%%%%%%%%%%%%%%%%%%%%%%%%%%

%%%%%%%%%%%%%%%%%%%%%%%%%%%%%%%%%%%%%%%%%%%%%%%%%%%%%%%%%%%%%%%%%%%%%%%%%%%%%%%%%%%%%%%%%%%%%%%%%%%%%%%
\section{Synthetic Personalization Task}\label{sec:synthetic}
We define the synthetic personalization environment as a contextual bandit setting consisting of a continuous state space $\mathcal{S} \subset \mathbb{R}^{d_\mathcal{S}}$, a discrete action space $\mathcal{A} \subset \mathbb{R}^{d_\mathcal{A} \times |\mathcal{A}|}$, and a reward function $r : \mathcal{S \times A} \to \mathcal{R}$ such that the value $r(s,a)$ depends on the latent state and action features that are not known to the agent.
The observed states are sampled from a distribution $\mathcal{T(S)}$ which is also unknown to the agent.

Similarly to a general contextual bandit task, the goal of an agent is to learn a policy $\pi$ that, for an observed state $s \in \mathcal{S}$, selects a relevant action $a_\pi \in \mathcal{A}$.
The relevance of the selected action $a_\pi$ for the given state $s$ is determined by the value of the reward $r(s,a_\pi) \in \mathbb{R}$.
The conventional way of training such an agent is to represent the agent as a neural network parameterized by the vector of weights $\theta$, and to allow the agent $\pi(\theta)$ to interact with the environment and adjust the policy's weights $\theta$ based on the received rewards.
For a general overview of the problem setting and common approaches, we refer the reader to the book~\cite{sutton2018reinforcement}.

\subsection{Synthetic Reward Signal}\label{sec:synthetic_reward}
\begin{figure}[!t]
    \centering
    \resizebox{\linewidth}{!}{
        \begin{tikzpicture}
            \node[draw, thick, circle]
                at (0,1) {$\ s \in \mathcal{S}\ $};
            \node[draw, thick, minimum height=9ex, minimum width=9em, text width=8em, text centered]
                at (3.5,1) {State Feature Extractor $\mathcal{F_S}$};
            \node[draw, thick, circle, minimum size=1cm]
                at (0,-1) {$\,a \in \mathcal{A}\ $};
            \node[draw, thick, minimum height=9ex, minimum width=9em, text width=8em, text centered]
                at (3.5,-1) {Action Feature Extractor $\mathcal{F_A}$};
            \node[draw, thick, minimum height=9ex, minimum width=9em, text width=8em, text centered]
                at (8,0) {Intrinsic Reward Function $\mathcal{F_L}$};
            \node[draw, thick, circle]
                at (11.5,0) {$r(s,a)$};
            \draw[-latex, thick] (.8,1) -- (1.8,1);
            \draw[-latex, thick] (.8,-1) -- (1.8,-1);
            \draw[-latex, thick] (5.2,1) -- (6.3,.2);
            \draw[-latex, thick] (5.2,-1) -- (6.3,-.2);
            \draw[-latex, thick] (9.7,0) -- (10.7,0);
        \end{tikzpicture}
    }
    \caption{A framework for generating synthetic reward functions.}
    \label{fig:reward}
\end{figure}
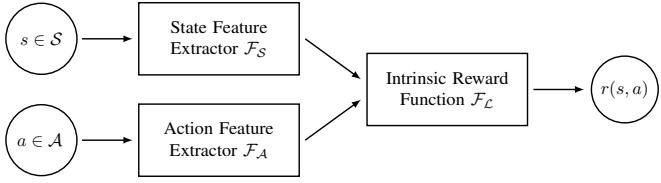

In order to simulate a complex yet realistic personalization task we assume that the reward signal received by the agent is determined by the state and action representations in the \textit{latent feature space} $\mathcal{L}$.
Accordingly, the constructed reward function consists of two \textit{latent feature extractors}~--- $\mathcal{F_S}$ for the state space and $\mathcal{F_A}$ for the action space~--- and an \textit{intrinsic reward function} $\mathcal{F_L}$ that determines the reward value based on the extracted features.
For the purpose of this paper the synthesized latent feature extractors $\mathcal{F_S}, \mathcal{F_A}$ consist of randomly generated neural networks of a particular structure described below, and the final reward value is determined by the proximity of the state and action feature vectors in the latent space $\mathcal{L}$, as displayed in Figure~\ref{fig:reward}.
Thus the synthetic reward function $r : \mathcal{S \times A \to R}$ consists of the following mappings:
\begin{itemize}
    \item \textit{State feature extractor} $\mathcal{F_S : S \to L}$,
    \item \textit{Action feature extractor} $\mathcal{F_A : A \to L}$,
    \item \textit{Intrinsic reward function} $\mathcal{F_L : L \times L \to R}$,
\end{itemize}
and the geometry of the reward function depends on the following parameters:
\begin{itemize}
    \item Dimensionality of the state and action spaces $\mathcal{S}$ and $\mathcal{A}$,
    \item Dimensionality of the latent feature space $\mathcal{L}$,
    \item Architecture of the state and action feature extractors,
    \item Structure of the intrinsic reward function.
\end{itemize}

Once the above parameters are specified, we generate a synthetic feature extractor as a feed-forward neural network with the input dimensionality $d_\mathcal{S} := \dim(\mathcal{S})$ (or $d_\mathcal{A} := \dim(\mathcal{A})$), output dimensionality $d_\mathcal{L} := \dim(\mathcal{L})$, and the selected architecture with the Gaussian activation functions $\sigma(z) = \exp(-z^2)$.
The network weights are initialized from the standard normal distribution $\mathcal{N}(0,1)$ so that the constructed mapping resembles a fully-trained network between the state space $\mathcal{S}$ (or the action space $\mathcal{A}$) and the latent feature space $\mathcal{L}$.
We then apply the \texttt{tanh} function to the output of both feature extractors $\mathcal{F_S}$ and $\mathcal{F_A}$ before computing the proximity in the latent feature space $\mathcal{L}$.
This concludes the construction of the synthetic feature extractors.

Lastly, we need to specify the intrinsic reward function $\mathcal{F_L} : \mathcal{L} \times \mathcal{L} \to \mathcal{R}$ that determines the reward value.
For simplicity, in this paper we consider the cosine similarity in $\ell_2$-norm as a measure of proximity of two feature vectors, i.e. our synthetic reward function $r : \mathcal{S \times A} \to \mathcal{R}$ has the form
\begin{equation}\label{eq:reward}
    r(s,a) = \mathcal{F_L} \big( \mathcal{F_S}(s), \mathcal{F_A}(a) \big)
    = \frac{\langle \mathcal{F_S}(s), \mathcal{F_A}(a) \rangle} {\|\mathcal{F_S}(s)\| \|\mathcal{F_A}(a)\|}
\end{equation}
where $\mathcal{F_S : S \to L}$ and $\mathcal{F_A : A \to L}$ are the synthetic state and action feature extractors respectively.

\subsection{Analysis of the Synthetic Reward Signal}
\begin{figure}[!t]
    \centering
    \includegraphics[width=\linewidth]{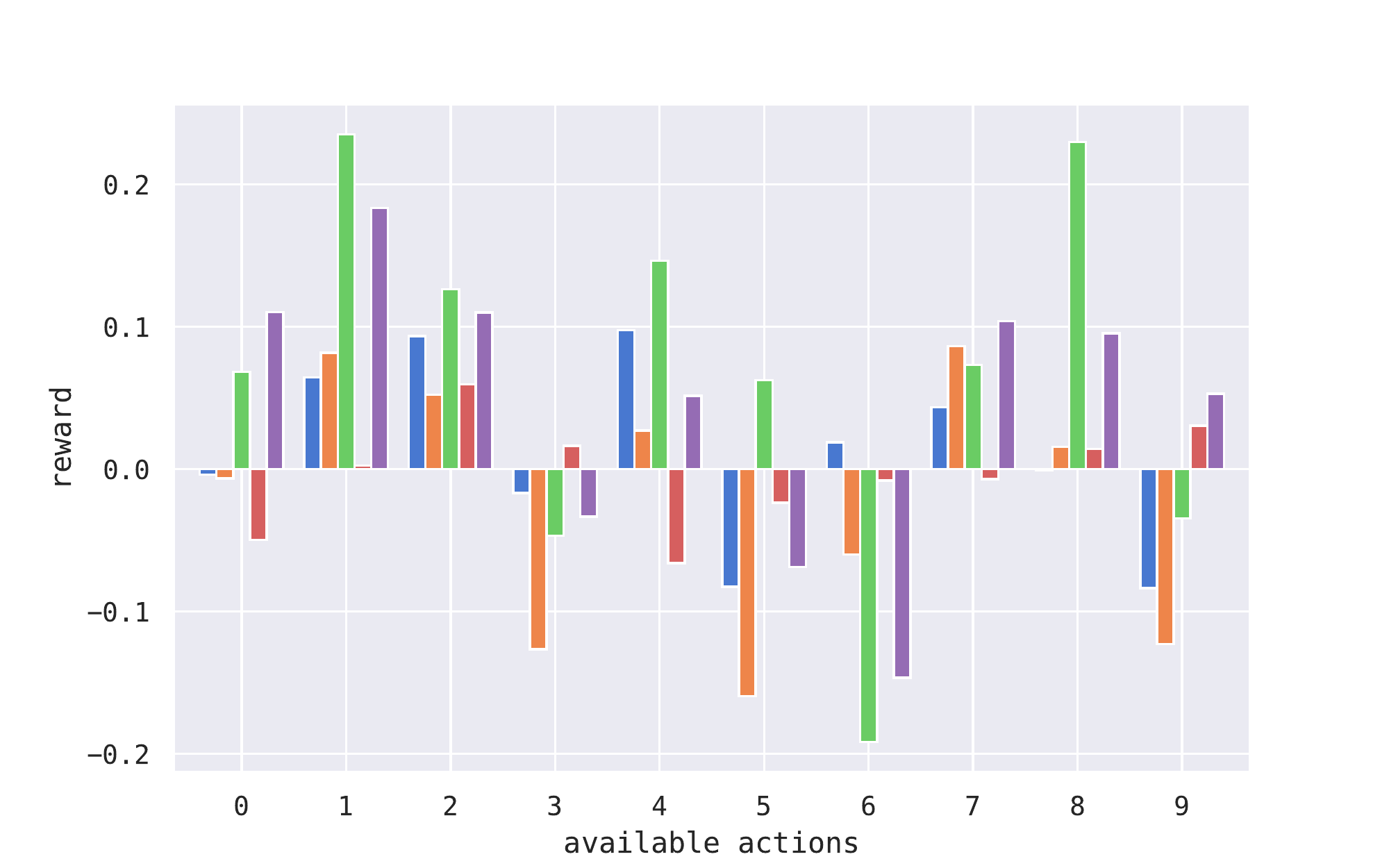}
    \caption{Rewards on the adjacent states in synthetic personalization task.}
    \label{fig:reward_distribution}
\end{figure}

In this section we take a closer look at the behavior of the synthetic reward function and demonstrate that, for a given state $s \in \mathcal{S}$, the reward values across the action space $\mathcal{A}$ are indeed determined by the latent features vector $\mathcal{F_S}(s) \in \mathcal{L}$ rather than by the raw coordinates in the state space.
Intuitively, this claim is demonstrated in Figure~\ref{fig:reward_distribution}, where we set up a synthetic personalization environment, sample states that are close in the state space, and report the corresponding reward values on the action space.
We observe that, despite being close in the state space, the reward values for the available actions are distinct, which demonstrates that the simulated reward signal depends on the latent features that an agent has to learn by interacting with the environment.
Below we provide the exact details on how the environment and reward signal are set up.

Specifically, we generate the synthetic environment with the state space $\mathcal{S} = [-1,1]^{100}$, the action space $\mathcal{A} = \{a_1, \ldots, a_{100}\}$ with $a_j \sim \mathcal{U}([-1,1]^{100})$, and the reward space $\mathcal{R} = [-1,1]$.
The synthetic reward function $r : \mathcal{S \times A \to R}$ is constructed according to~\eqref{eq:reward}, where the state and action feature extractors $\mathcal{F_S}$ and $\mathcal{F_A}$ are the randomly generated feed-forward neural networks with the architecture $[100,100,100]$, as described in Section~\ref{sec:synthetic_reward}, i.e.
\begin{align*}
    \mathcal{F_S}(s) = \texttt{tanh(FC100(FC100(FC100($s$))))},
    \\
    \mathcal{F_A}(a) = \texttt{tanh(FC100(FC100(FC100($a$))))},
\end{align*}
where \texttt{FC100} denotes a fully-connected layer with $100$ nodes and the Gaussian activation function $\sigma(z) = \exp(-z^2)$, with the weights sampled from the standard normal distribution $\mathcal{N}(0,1)$.
The output of each feature extractor is a vector in the latent feature space $\mathcal{L} = [-\nicefrac{\pi}{2},\nicefrac{\pi}{2}]^{100}$, and the value of the reward $r(s,a)$ is the cosine similarity between the feature vectors $\mathcal{F_S}(s) \in \mathcal{L}$ and $\mathcal{F_A}(a) \in \mathcal{L}$.

Figure~\ref{fig:reward_distribution} is obtained by considering the set of states $\{s_1, \ldots, s_5\} \in \mathcal{S}$ whose coordinates are sampled from the normal distribution $\mathcal{N}(0,0.0001)$ and computing the corresponding reward values on the first ten available actions $a_0, \ldots, a_9 \in \mathcal{A}$.
We note that, even though the sampled states $\{s_1, \ldots, s_5\}$ are close in the state space $\mathcal{S}$, the distribution of the rewards values $\{r(s_1,a_j), \ldots, r(s_5,a_j)\}$ is not similar for any $0 \le j \le 9$.

\begin{figure}[!t]
    \centering
    \includegraphics[width=\linewidth]{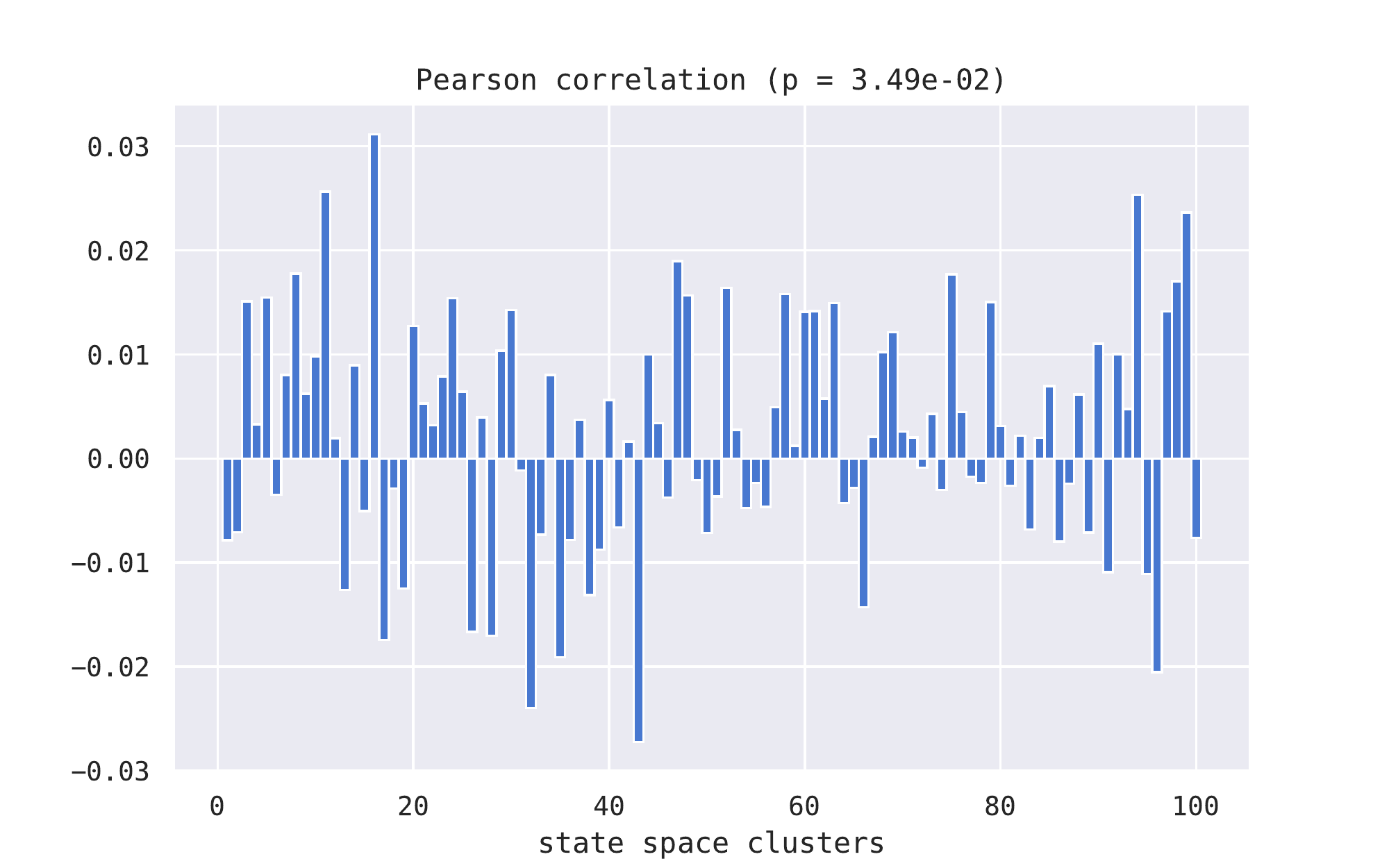}
    \caption{Correlation between the state clustering and reward values.}
    \label{fig:reward_correlation}
\end{figure}

For a more technical demonstration of the relation between the proximity in the state space and the distribution of the rewards across the action space, we uniformly sample $100,000$ states $\{s_i\}_{i=1}^{100,000}$ from the state space $\mathcal{S}$ and then split them into $100$ clusters $C_1, \ldots, C_{100}$ via the K-means algorithm.
For each cluster $C_k$ we compute the vector of reward values $r_\mathcal{A}(s) := \{r(s,a_0), \ldots, r(s,a_{99})\} \in \mathbb{R}^{100}$ for every state $s \in C_k$.
Then we calculate the Pearson's correlation coefficient between the set of pair-wise differences of the states in the cluster and the corresponding vectors of rewards, i.e.
\begin{multline*}
    \rho_k := \textrm{corr}\Big(
    \big\{ \big\| s - s' \big\| \,\big|\, s,s' \in C_k \big\},
    \\
    \big\{ \big\| r_\mathcal{A}(s) - r_\mathcal{A}(s') \big\| \,\big|\, s,s' \in C_k \big\}
    \Big)
\end{multline*}
for each cluster $C_k$, $1 \le k \le 100$.
The resulting correlation coefficients are presented in Figure~\ref{fig:reward_correlation}.
Once again, we observe no correlation between the distance in the state space $\mathcal{S}$ and the distance in the reward space $\mathcal{R}$.

Given these empirical demonstrations and the overall intuition behind the construction of a synthetic personalization task, one would assume that clustering of the state space with the intent of simplifying the task is a counter-productive effort.
However, our extensive numerical results, discussed in the next section, quite unexpectedly show the opposite.
%%%%%%%%%%%%%%%%%%%%%%%%%%%%%%%%%%%%%%%%%%%%%%%%%%%%%%%%%%%%%%%%%%%%%%%%%%%%%%%%%%%%%%%%%%%%%%%%%%%%%%%

%%%%%%%%%%%%%%%%%%%%%%%%%%%%%%%%%%%%%%%%%%%%%%%%%%%%%%%%%%%%%%%%%%%%%%%%%%%%%%%%%%%%%%%%%%%%%%%%%%%%%%%
\section{Numerical Experiments}\label{sec:numerics}
In this section we train agents with several commonly-used RL algorithms on the Synthetic Personalization Tasks defined in Section~\ref{sec:synthetic}, and compare their performance.
Specifically, we consider the following algorithms:
\begin{itemize}
    \item \texttt{A2C}~--- Advantage Actor Critic~\cite{mnih2016asynchronous};
    \item \texttt{DQN}~--- Deep Q Network~\cite{mnih2013playing};
    \item \texttt{PPO}~--- Proximal Policy Optimization~\cite{schulman2017proximal}.
\end{itemize}
The presented RL algorithms are implemented through the use of the \texttt{Stable-Baselines3}\footnote{\url{https://stable-baselines3.readthedocs.io/}} library with the recommended choice of hyperparameters.
Our numerical experiments are performed in Python~3.8 and the source code reproducing the presented results 
%will be made available at the time of publication.
is publicly available at~\url{https://github.com/sukiboo/hyperpersonalization}.

While the conventional way of assessing an agent's performance on a contextual bandit problem is to measure the total regret, it is not the most suitable metric in our setting.
First, in real-world personalization applications it is often not feasible to fully learn the environment due to the complexity of the reward signal, regardless of whether it is caused by the nontrivial geometry of the state-action space or by external factors, such as the stochasticity or nonstationarity of the environment.
Second, it is important to ensure that the trained policy provides a nontrivial level of performance shortly after being deployed, i.e. with a limited number of agent-environment interactions.
What qualifies a nontrivial performance level is typically application-specific but the common benchmark in personalization tasks is the expected return of the uniformly random agent.
Hence if an agent's return is consistently higher that that of the uniformly random agent, we say that a nontrivial level of performance is achieved, which is a necessary condition for deploying an agent.

To this end, in our numerical experiments the performance of an agent is measured and reported to reflect the above points.
Namely, for each instance of the synthetic environment we obtain the uniformly random policy $\pi_0$ and the optimal policy $\pi^\star$, i.e. such that for any state $s \in \mathcal{S}$ we have
\begin{align*}
    \pi_0(s) &= \mathbb{E} \big[ r(s,a) \,\big|\, a \sim \mathcal{U(A)} \big]
    = \frac{1}{|\mathcal{A}|} \sum_{a \in \mathcal{A}} r(s,a),
    \\
    \pi^\star(s) &= \max_{a \in \mathcal{A}} r(s,a).
\end{align*}
Then the performance of the target policy $\pi$ on a state $s \in \mathcal{S}$ is reported as
\begin{equation}\label{eq:perf}
    R(\pi | s) = \frac{\pi(s) - \pi_0(s)}{\pi^\star(s) - \pi_0(s)}.
\end{equation}
The positive value of $R(\pi|\cdot)$ indicates that the agent outperforms the uniformly random policy on the given state, and the value of $1$ means that the optimal action is selected for the given state.

\subsection*{Experiment Setup}
\begin{figure}[!t]
    \centering
    \includegraphics[width=\linewidth]{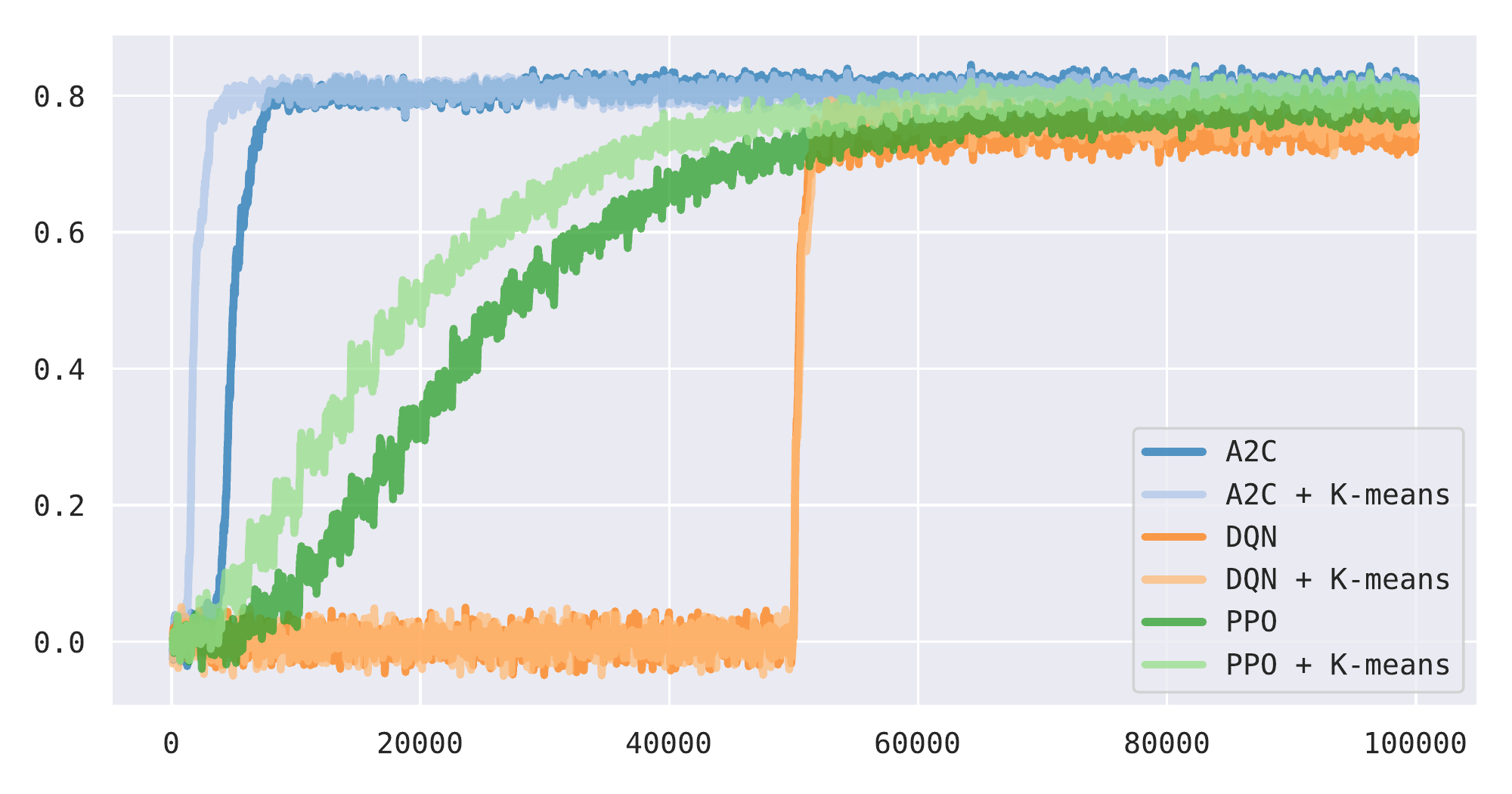}
    \caption{Low-dimensional environment, simple reward, large networks.}
    \label{fig:ex1}
\end{figure}

\begin{figure}[!t]
    \centering
    \includegraphics[width=\linewidth]{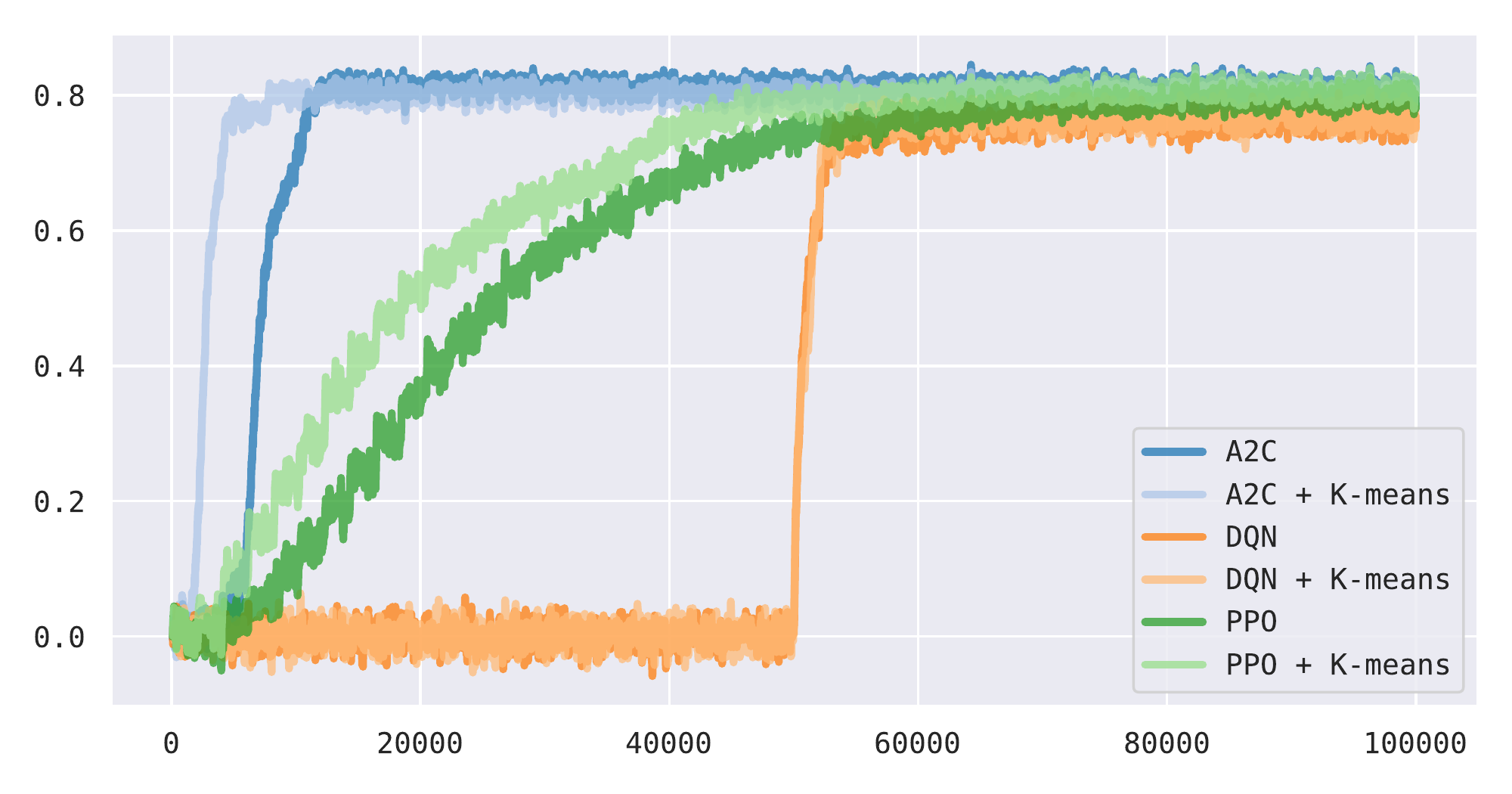}
    \caption{Low-dimensional environment, simple reward, medium networks.}
    \label{fig:ex2}
\end{figure}

\begin{figure}[!t]
    \centering
    \includegraphics[width=\linewidth]{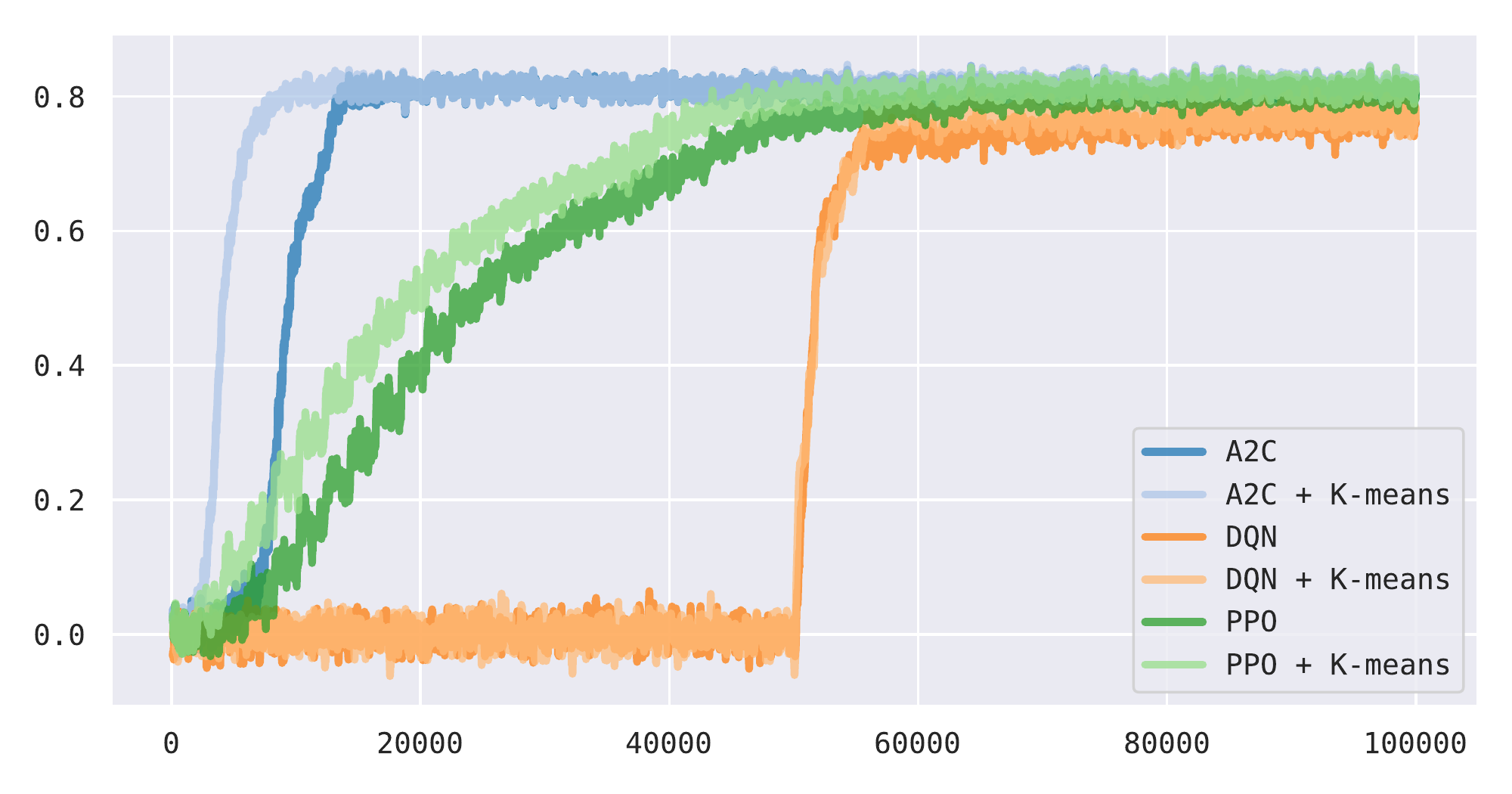}
    \caption{Low-dimensional environment, simple reward, small networks.}
    \label{fig:ex3}
\end{figure}

\begin{figure}[!t]
    \centering
    \includegraphics[width=\linewidth]{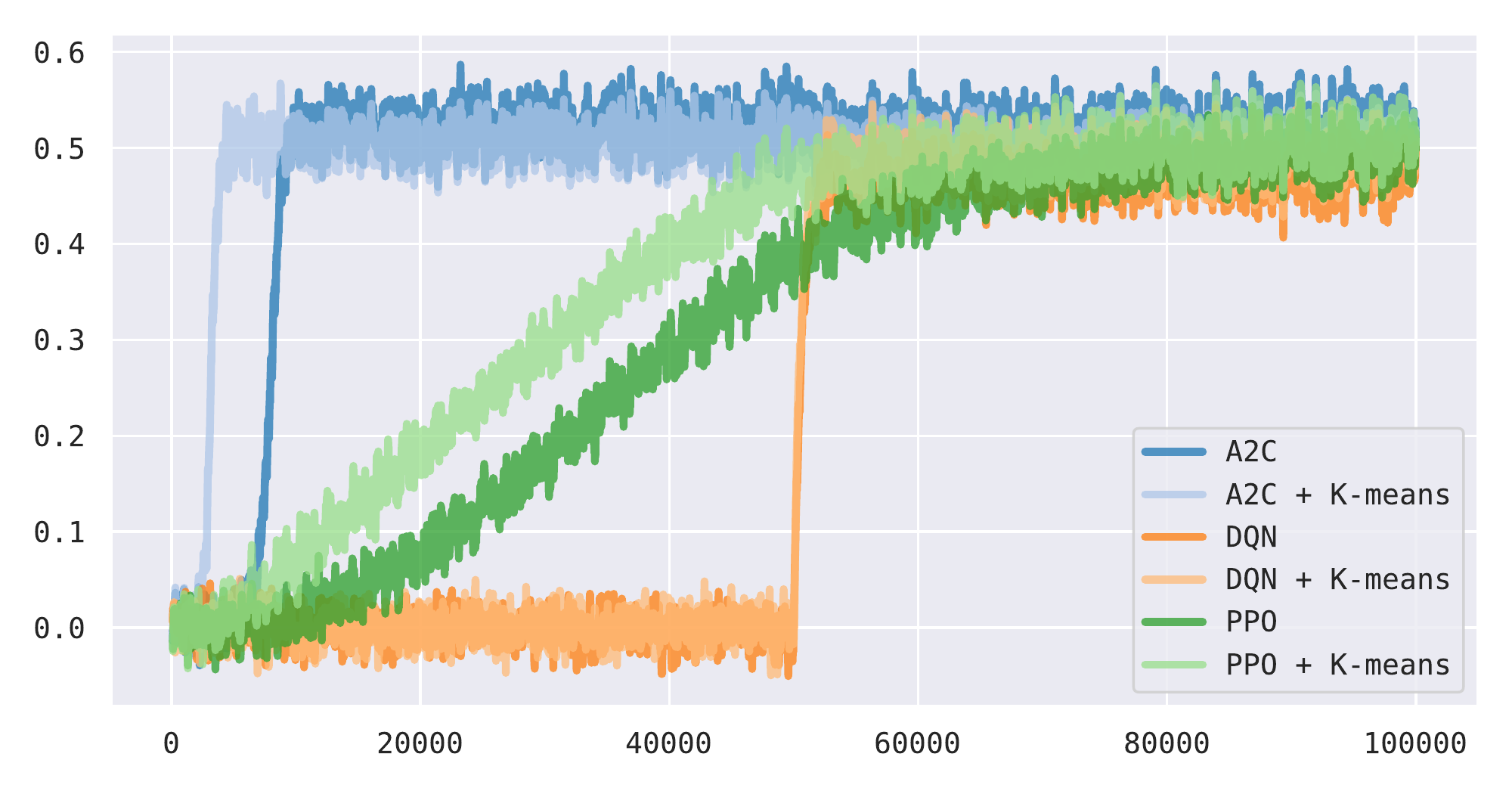}
    \caption{Low-dimensional environment, complex reward, large networks.}
    \label{fig:ex4}
\end{figure}

\begin{figure}[!t]
    \centering
    \includegraphics[width=\linewidth]{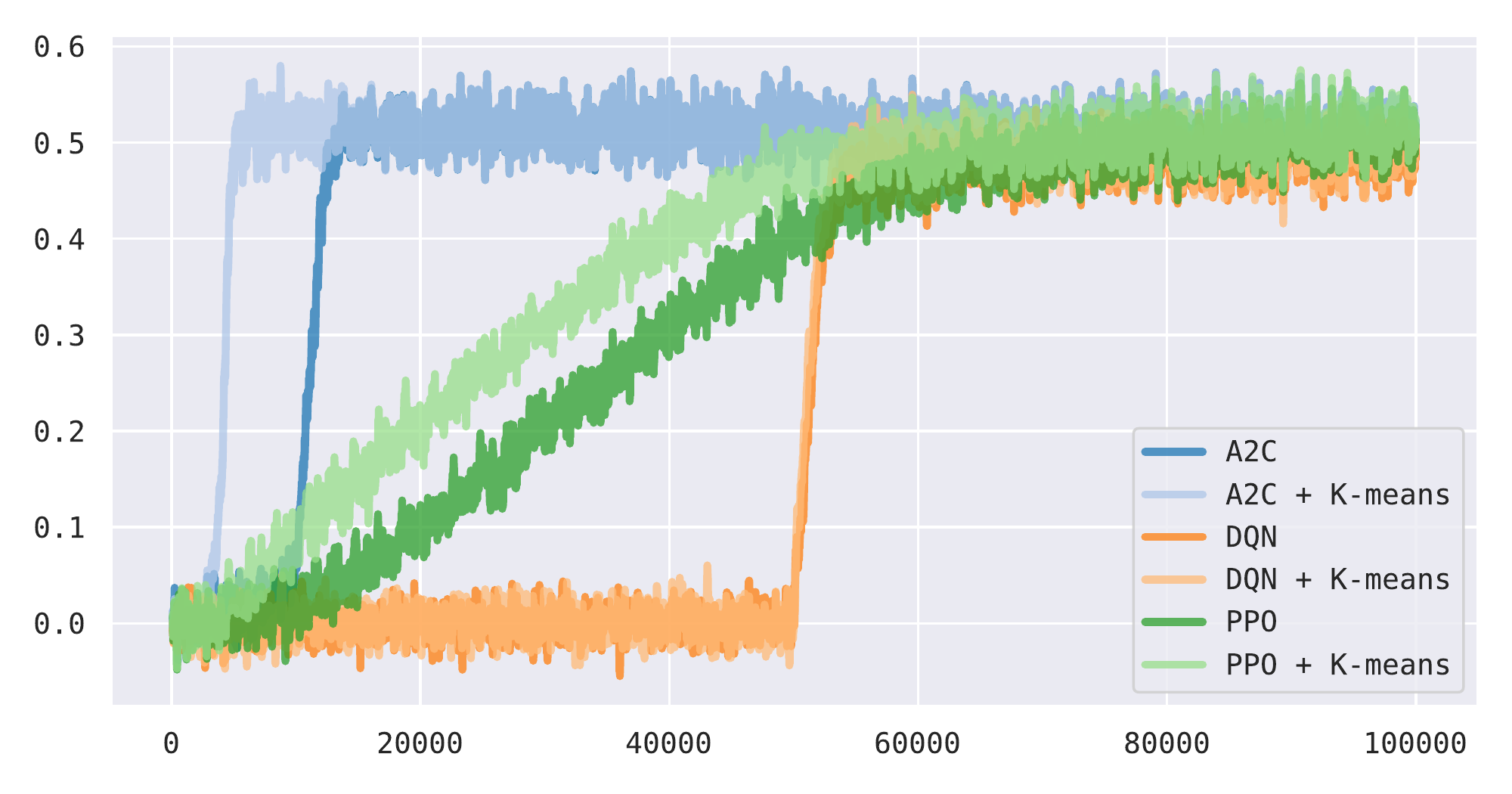}
    \caption{Low-dimensional environment, complex reward, medium networks.}
    \label{fig:ex5}
\end{figure}

\begin{figure}[!t]
    \centering
    \includegraphics[width=\linewidth]{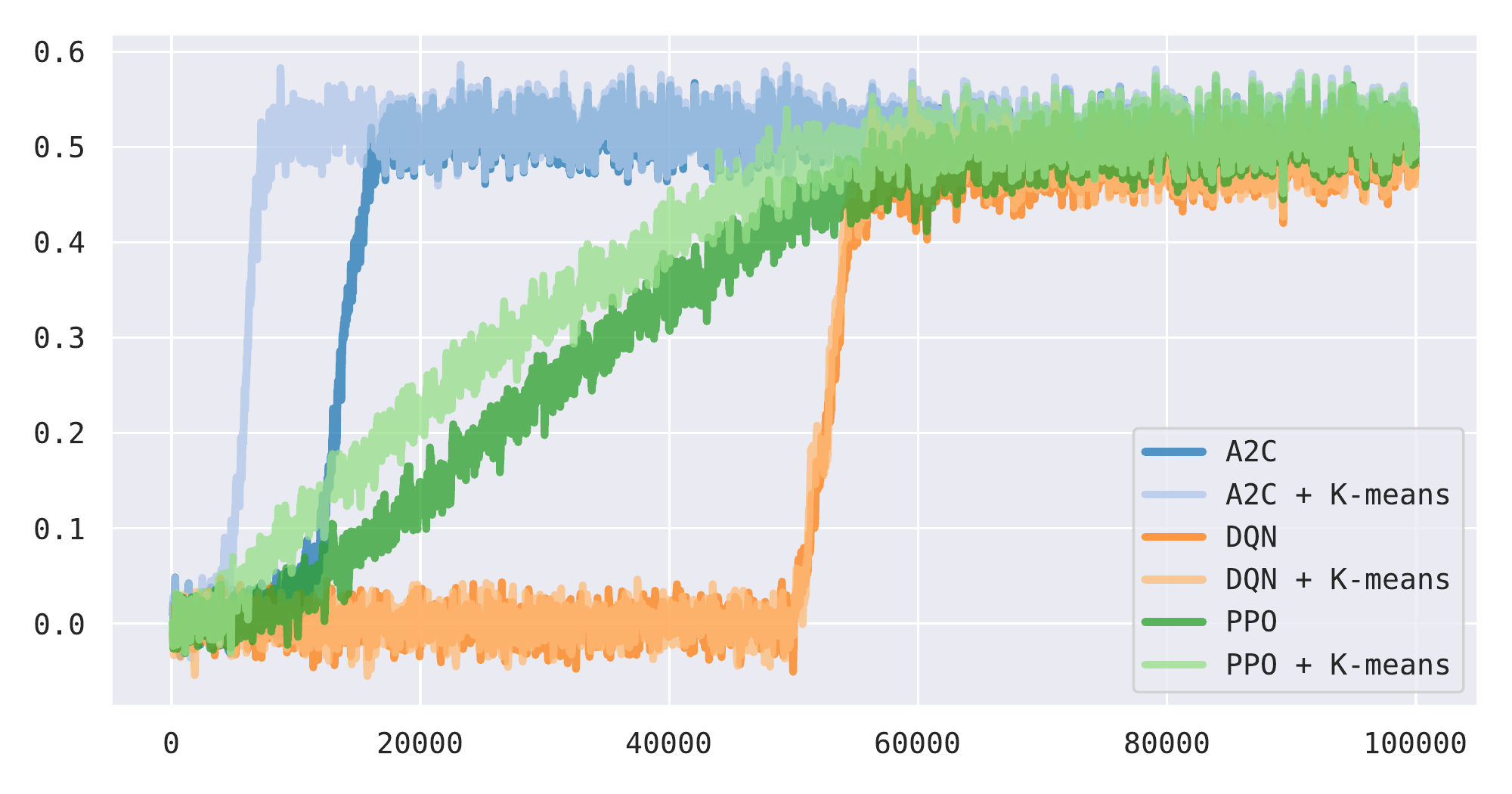}
    \caption{Low-dimensional environment, complex reward, small networks.}
    \label{fig:ex6}
\end{figure}

\begin{figure}[!t]
    \centering
    \includegraphics[width=\linewidth]{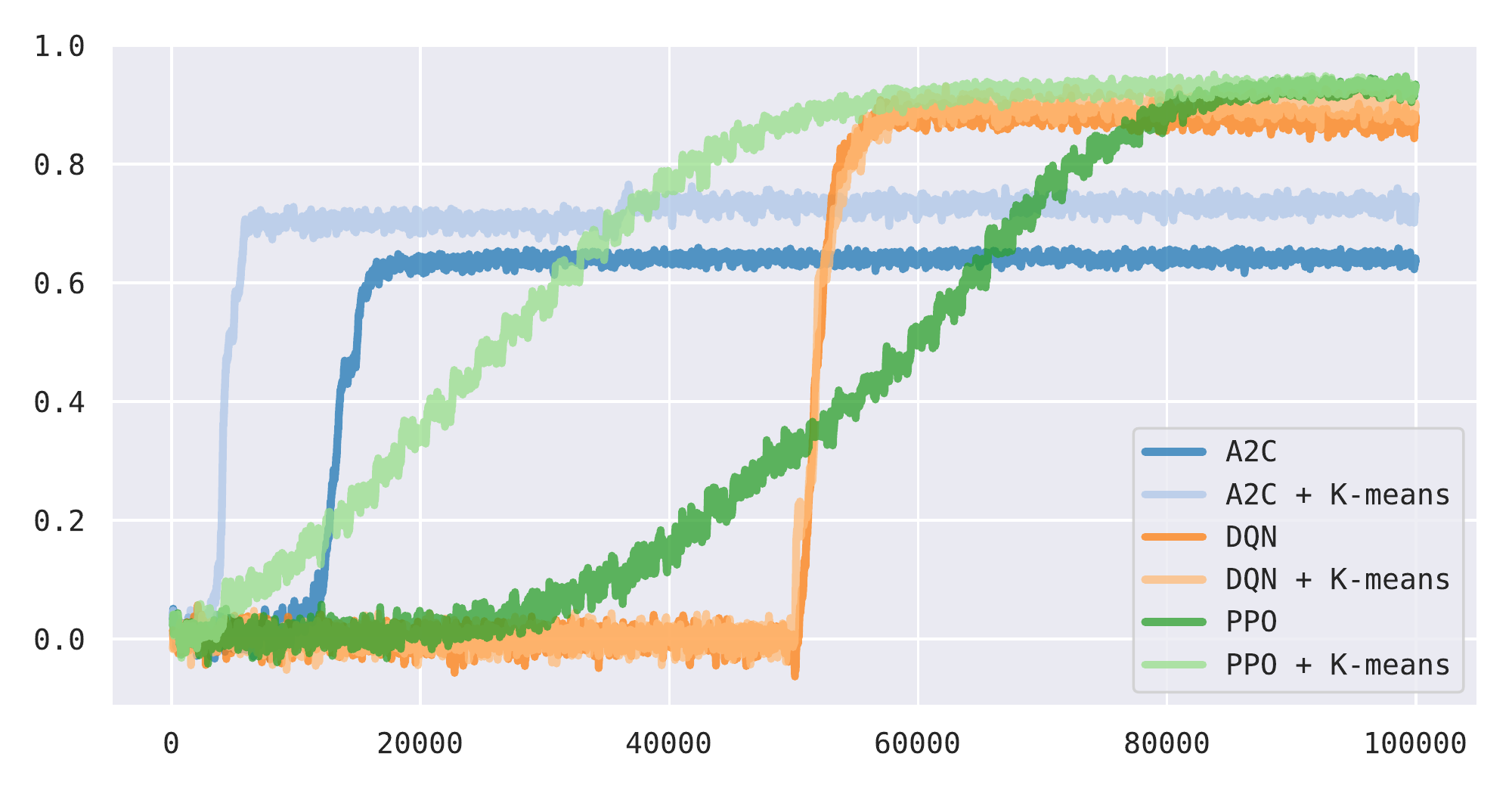}
    \caption{High-dimensional environment, simple reward, large networks.}
    \label{fig:ex7}
\end{figure}

\begin{figure}[!t]
    \centering
    \includegraphics[width=\linewidth]{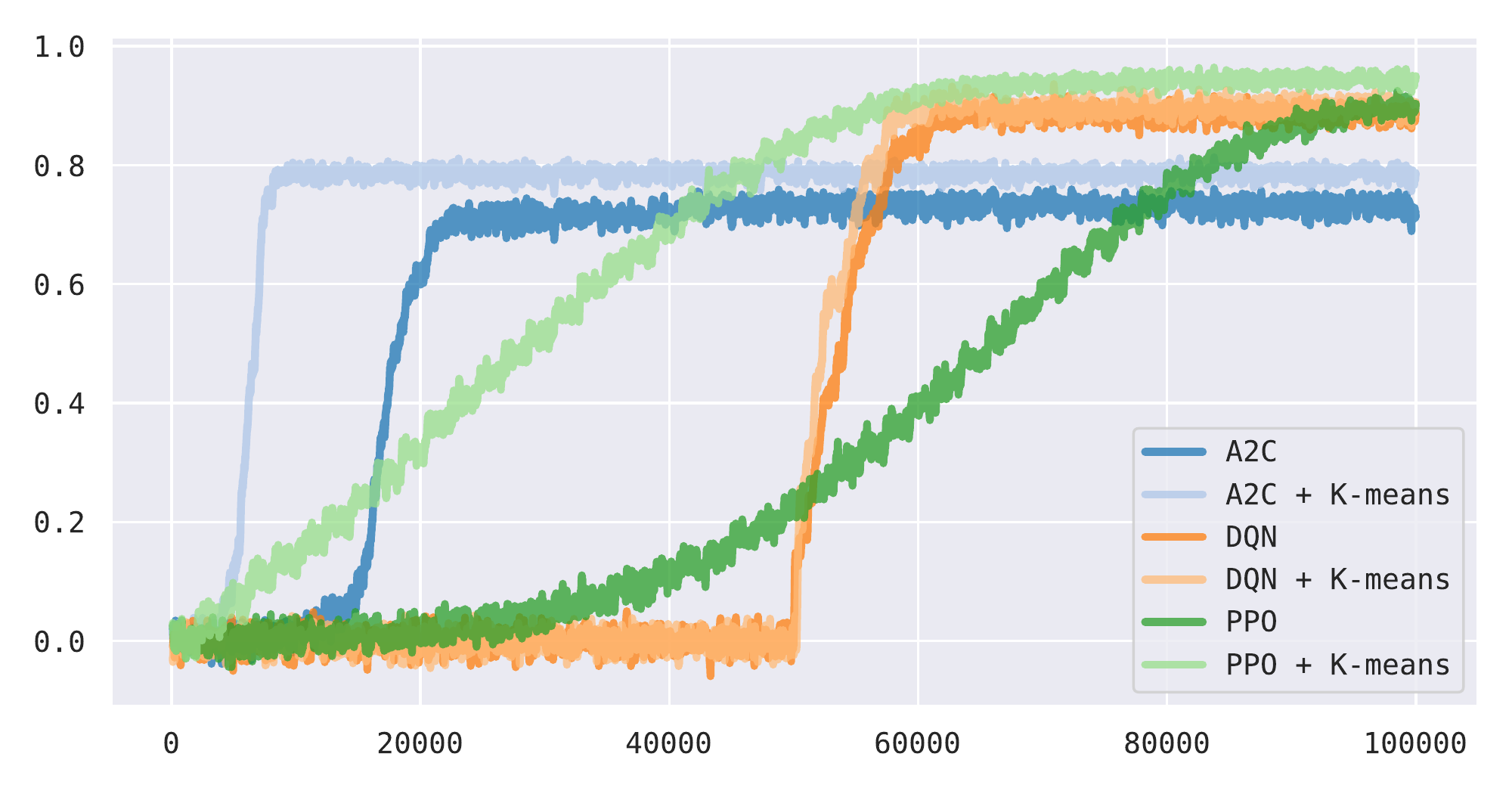}
    \caption{High-dimensional environment, simple reward, medium networks.}
    \label{fig:ex8}
\end{figure}

\begin{figure}[!t]
    \centering
    \includegraphics[width=\linewidth]{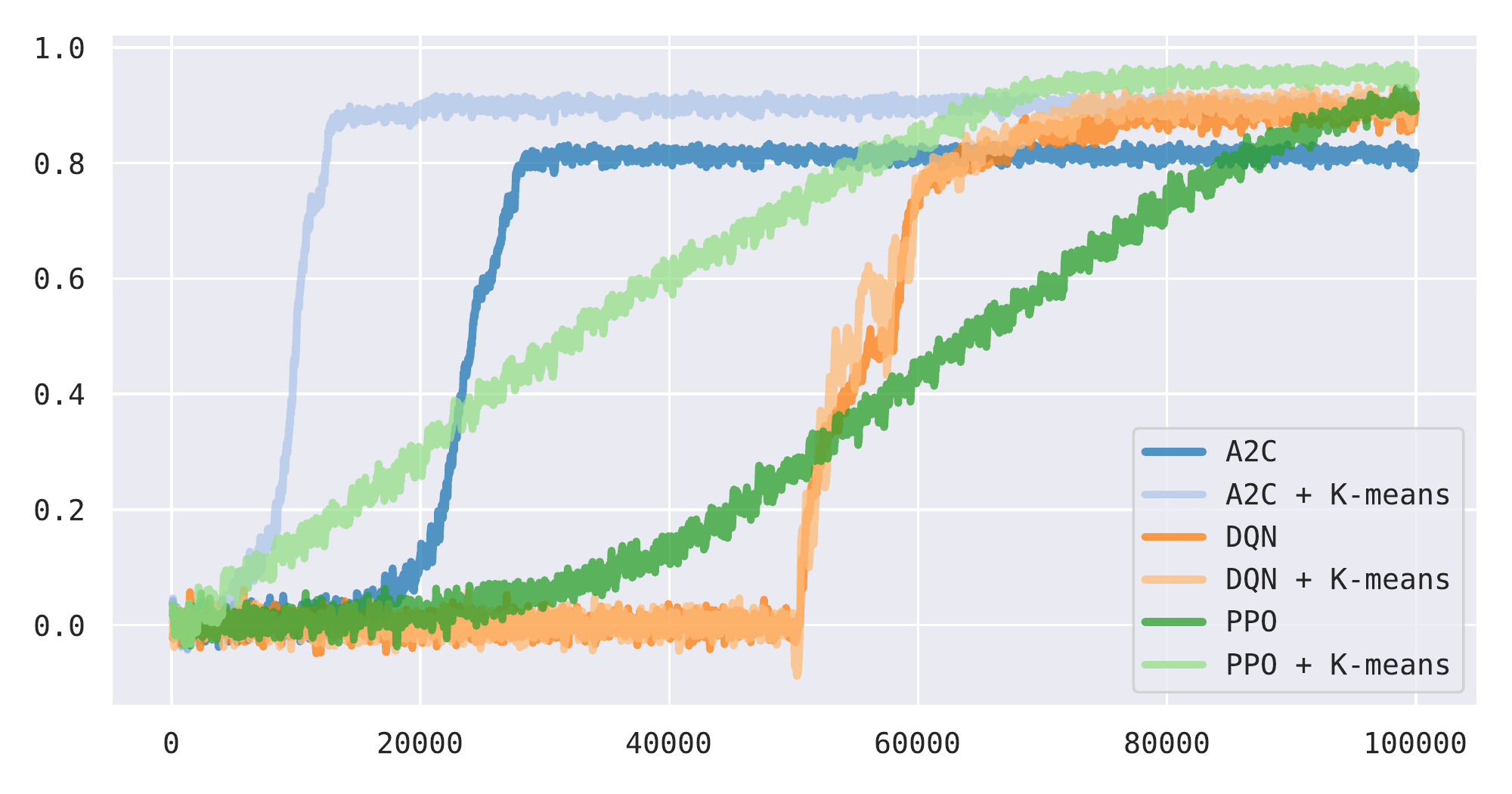}
    \caption{High-dimensional environment, simple reward, small networks.}
    \label{fig:ex9}
\end{figure}

\begin{figure}[!t]
    \centering
    \includegraphics[width=\linewidth]{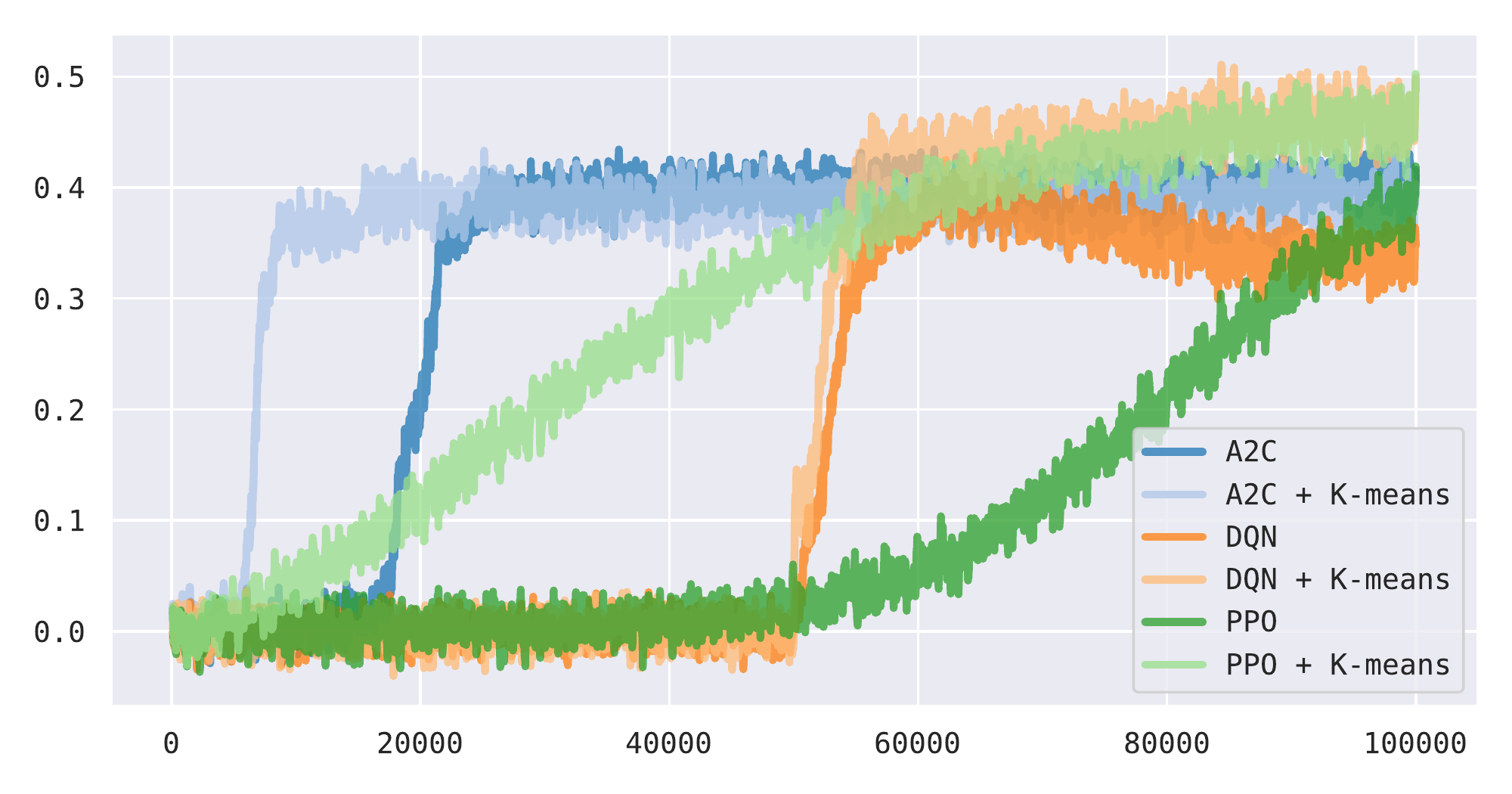}
    \caption{High-dimensional environment, complex reward, large networks.}
    \label{fig:ex10}
\end{figure}

\begin{figure}[!t]
    \centering
    \includegraphics[width=\linewidth]{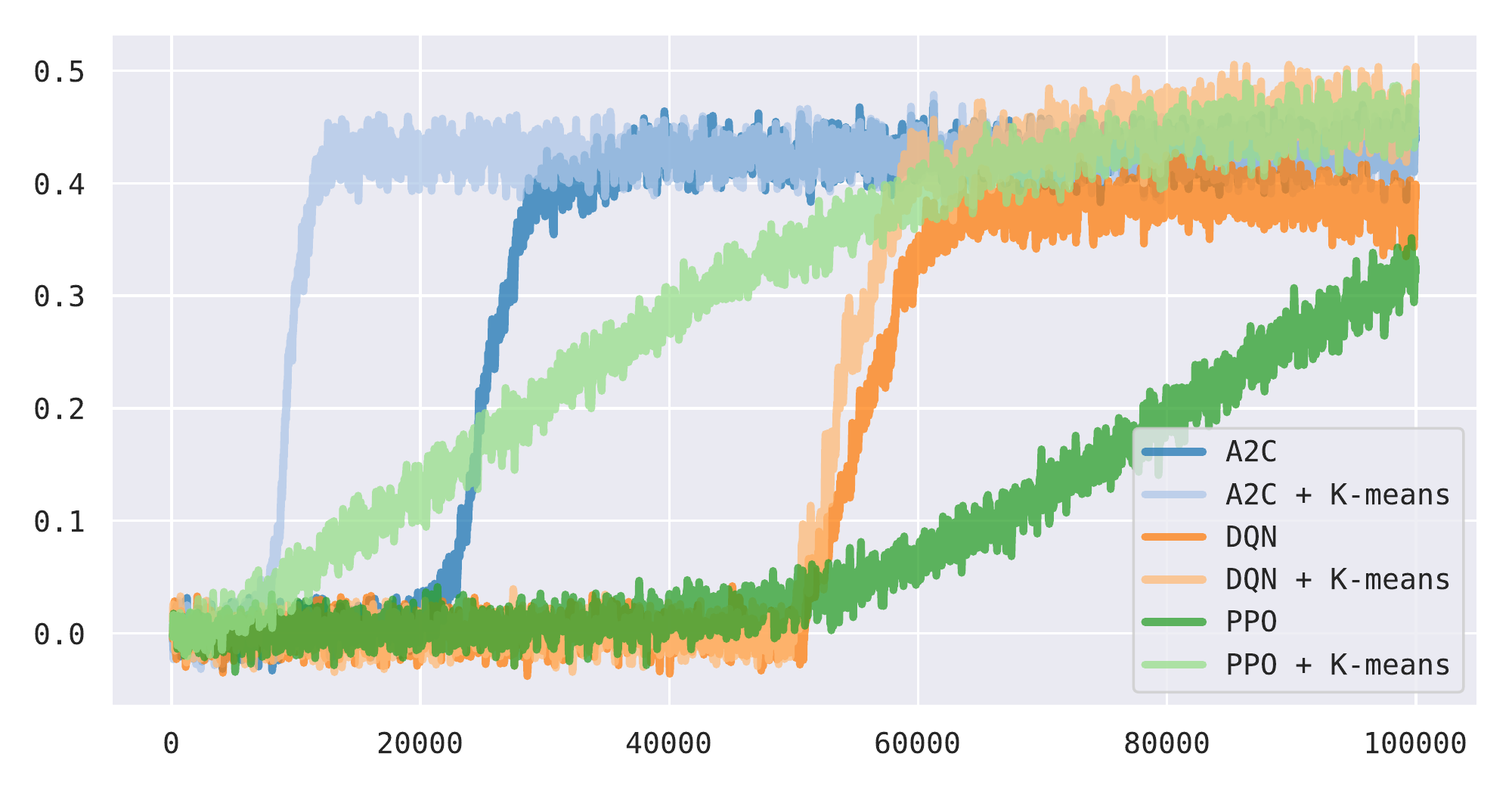}
    \caption{High-dimensional environment, complex reward, medium networks.}
    \label{fig:ex11}
\end{figure}

\begin{figure}[!t]
    \centering
    \includegraphics[width=\linewidth]{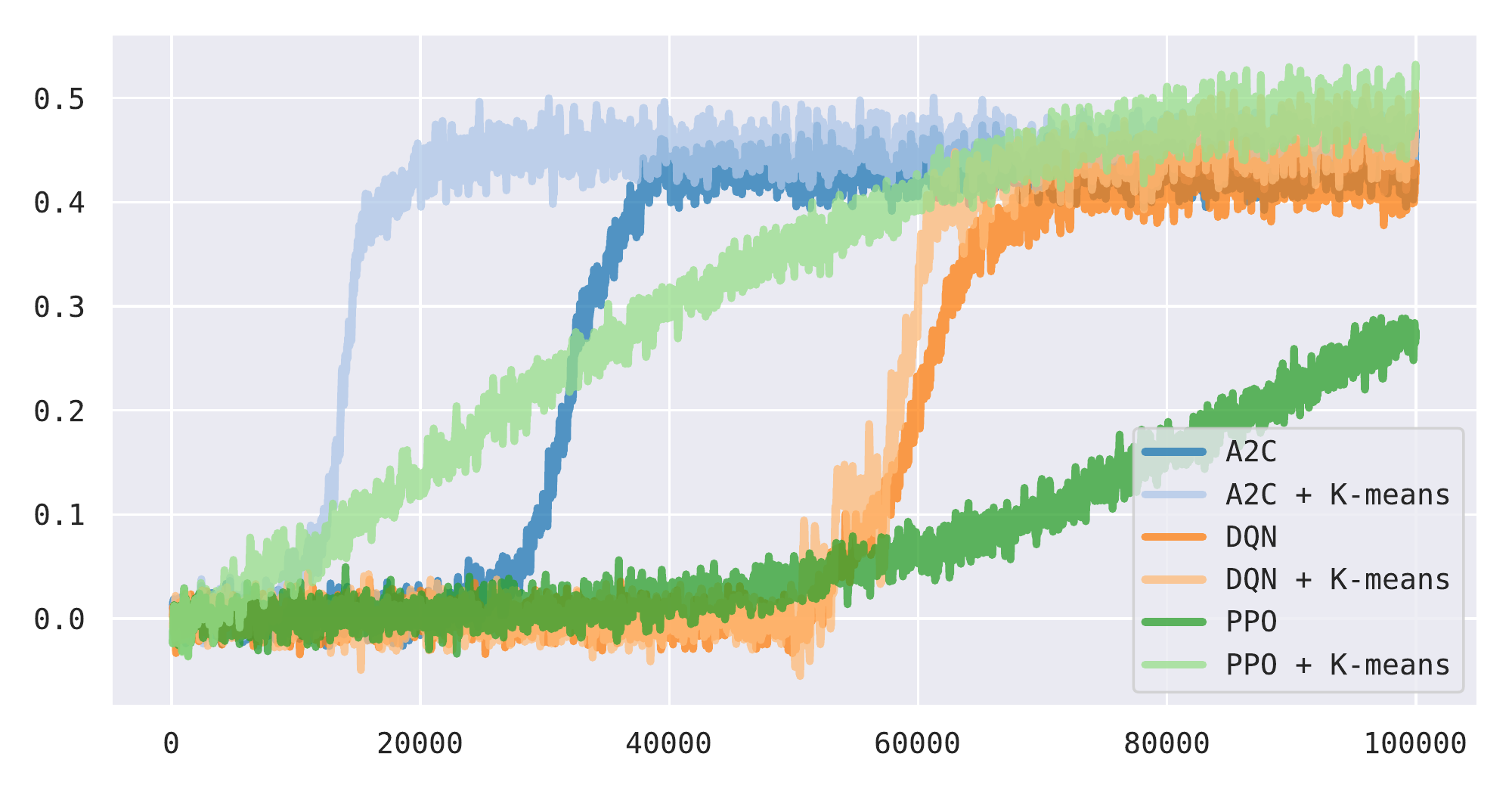}
    \caption{High-dimensional environment, complex reward, small networks.}
    \label{fig:ex12}
\end{figure}

The process of generating the synthetic environment and reward signal allows for a lot of freedom in the selection of various parameters.
In order to provide a wide range of experiments, we vary the complexity of different components, listed in Table~\ref{tab:params}, and report the results in Figures~\ref{fig:ex1}--\ref{fig:ex12}.
Specifically, we distinguish three components that contribute to the overall complexity of the Synthetic Personalization Task: the dimensionality of the environment, the geometry of the reward signal, and the architecture of the policy's network.

\begin{table*}[!t]
    \caption{Experiment Parameters}
    \label{tab:params}
    \centering
    \begin{tabular}{lcccccc}
        \toprule
        & $\operatorname{card}(\mathcal{A})$ & $d_{\mathcal{S}}$ & $d_{\mathcal{A}}$ & $d_{\mathcal{L}}$ & $r$-architecture & $\pi$-architecture
        \\\midrule
        Figure~\ref{fig:ex1} & 100 & 100 & 100 & 10 & [10,10,10] & [128,128,128]
        \\
        Figure~\ref{fig:ex2} & 100 & 100 & 100 & 10 & [10,10,10] & [64,64,64]
        \\
        Figure~\ref{fig:ex3} & 100 & 100 & 100 & 10 & [10,10,10] & [32,32,32]
        \\
        Figure~\ref{fig:ex4} & 100 & 100 & 100 & 100 & [100,100,100] & [128,128,128]
        \\
        Figure~\ref{fig:ex5} & 100 & 100 & 100 & 100 & [100,100,100] & [64,64,64]
        \\
        Figure~\ref{fig:ex6} & 100 & 100 & 100 & 100 & [100,100,100] & [32,32,32]
        \\
        Figure~\ref{fig:ex7} & 1,000 & 1,000 & 1,000 & 10 & [10,10,10] & [128,128,128]
        \\
        Figure~\ref{fig:ex8} & 1,000 & 1,000 & 1,000 & 10 & [10,10,10] & [64,64,64]
        \\
        Figure~\ref{fig:ex9} & 1,000 & 1,000 & 1,000 & 10 & [10,10,10] & [32,32,32]
        \\
        Figure~\ref{fig:ex10} & 1,000 & 1,000 & 1,000 & 100 & [100,100,100] & [128,128,128]
        \\
        Figure~\ref{fig:ex11} & 1,000 & 1,000 & 1,000 & 100 & [100,100,100] & [64,64,64]
        \\
        Figure~\ref{fig:ex12} & 1,000 & 1,000 & 1,000 & 100 & [100,100,100] & [32,32,32]
        \\\bottomrule
    \end{tabular}
\end{table*}

As described in Section~\ref{sec:synthetic}, the synthetic environment consists of the continuous state space $\mathcal{S} \subset \mathbb{R}^{d_\mathcal{S}}$ and the discrete action space $\mathcal{A} \subset \mathbb{R}^{d_\mathcal{A} \times |\mathcal{A}|}$.
For the purpose of this work the state space is chosen to be the hypercube $[-1,1]^{d_\mathcal{S}}$ and the distribution of the observed states is uniform, i.e.
\[
    \mathcal{S} = [-1,1]^{d_\mathcal{S}}
    \ \text{ and }\ 
    \mathcal{T(S) = U(S)},
\]
and thus
\[
    s \sim \mathcal{T(S)} = \mathcal{U}([-1,1]^{d_\mathcal{S}}).
\]
Similarly, the action space is constructed by uniformly sampling the hypercube $[-1,1]^{d_\mathcal{A}}$, i.e.
\[
    \mathcal{A} = \{a_1, \ldots, a_{|\mathcal{A}|}\},
    \text{ where }\ 
    a_j \sim \mathcal{U}([-1,1]^{d_\mathcal{A}}).
\]
As for the value of the parameters $d_\mathcal{S}, d_\mathcal{A}$, and $|\mathcal{A}|$, we consider two cases:
\begin{itemize}
    \item \textit{Low-dimensional}: $|\mathcal{A}| = d_\mathcal{S} = d_\mathcal{A} = 100$;
    \item \textit{High-dimensional}: $|\mathcal{A}| = d_\mathcal{S} = d_\mathcal{A} = 1,000$.
\end{itemize}

The synthetic reward signal $r : \mathcal{S \times A} \to \mathcal{R}$ is generated according to the process described in Section~\ref{sec:synthetic_reward}.
The complexity of the reward function is determined by the dimensionality $d_\mathcal{L}$ of the latent feature space $\mathcal{L}$ and the architecture of the state and action feature extractors.
For the ease of notation, we employ the same architecture for the state and action branches, denoted as $r$-architecture.
We consider two cases for the geometry of the reward signal:
\begin{itemize}
    \item \textit{Simple}: $d_\mathcal{L} = 10$ and $r$-architecture is $[10,10,10]$;
    \item \textit{Complex}: $d_\mathcal{L} = 100$ and $r$-architecture is $[100,100,100]$.
\end{itemize}

Finally, the architecture of the neural network that is used to parameterize the agent's policy is explicitly related to the agent's ability to learn the environment.
Hence we specify three choices for the network architecture, denoted as $\pi$-architecture:
\begin{itemize}
    \item \textit{Large}: $\pi$-architecture is $[128,128,128]$;
    \item \textit{Medium}: $\pi$-architecture is $[64,64,64]$;
    \item \textit{Small}: $\pi$-architecture is $[32,32,32]$.
\end{itemize}
Note that some of the algorithms we employ (namely, \texttt{A2C} and \texttt{PPO}) use two networks: one for the actor and one for the critic.
In our experiments both networks use the same architecture.


The exact parameter values for each experiment are outlined in Table~\ref{tab:params}.
Once the parameters are chosen, a single experiment consists of the following steps:
\begin{enumerate}
    \item Generate the environment;
    \item Synthesize the reward function;
    \item Train RL agents on the environment;
    \item Cluster the state space via the K-means;
    \item Train RL agents on the clustered environment;
    \item Report the agents' performance.
\end{enumerate}
Each agent is deployed for $100,000$ timesteps for training with the corresponding approach, i.e. each algorithm has access to $100,000$ unique agent-environment interactions, not accounting for samples from the replay buffer.
Since the performance of the agent might be sensitive to the initialization of the network's weights, each agent is trained $3$ times and the average result is reported.
We also note that, despite being completely synthetic, our setting is completely reproducible and each agent is observing exactly the same sequence of states, regardless of the stochasticity in the corresponding RL algorithm.
In order to avoid a possible unintentional bias from the synthetically constructed environment, each experiment is repeated $3$ times with an independently generated state space, action space, and reward function, and the average of the normalized returns, computed via~\eqref{eq:perf}, are reported in Figures~\ref{fig:ex1}--\ref{fig:ex12}.
We note that while the presented experiments only use the parameter values listed in Table~\ref{tab:params}, a more extensive range of experiments on other configurations has provided similar results.
%%%%%%%%%%%%%%%%%%%%%%%%%%%%%%%%%%%%%%%%%%%%%%%%%%%%%%%%%%%%%%%%%%%%%%%%%%%%%%%%%%%%%%%%%%%%%%%%%%%%%%%

%%%%%%%%%%%%%%%%%%%%%%%%%%%%%%%%%%%%%%%%%%%%%%%%%%%%%%%%%%%%%%%%%%%%%%%%%%%%%%%%%%%%%%%%%%%%%%%%%%%%%%%
\section{Conclusion}\label{sec:conclusion}
In this effort we consider personalization tasks and discuss some of the common RL-based approaches and the practical techniques used to decrease the dimensionality of the problem, typically achieved via some form of clustering.
In order to empirically evaluate the efficiency of different approaches, we develop a framework for generating synthetic personalization environments with a reward signal of arbitrary complexity, which can be used to simulate a wide range of real-world applications.

We argue that in such complicated environments most of the basic forms of clustering should not work, since naively restricting the dimensionality of the environment representation unavoidably leads to the loss of information.
However, seemingly counter-intuitively, we observe that, when the conventional RL algorithms are deployed, even the most trivial form of clustering accelerates learning and does not restrict the performance.
We hypothesize that this phenomenon is due to the employment of conventional RL algorithms that by design are not able to capture the complicated mechanisms behind the composition of the reward signals that are common in real-world applications.
As a particular observation, our results imply that the information lost during the clustering process is not used by the conventional agents, which calls for the development of more sophisticated approaches, which is the topic of our future work.
%%%%%%%%%%%%%%%%%%%%%%%%%%%%%%%%%%%%%%%%%%%%%%%%%%%%%%%%%%%%%%%%%%%%%%%%%%%%%%%%%%%%%%%%%%%%%%%%%%%%%%%

%%%%%%%%%%%%%%%%%%%%%%%%%%%%%%%%%%%%%%%%%%%%%%%%%%%%%%%%%%%%%%%%%%%%%%%%%%%%%%%%%%%%%%%%%%%%%%%%%%%%%%%
\bibliographystyle{abbrv}
\bibliography{references}

\begin{thebibliography}{10}

\bibitem{Abel-16}
D.~Abel, D.~Hershkowitz, and M.~Littman.
\newblock Near optimal behavior via approximate state abstraction.
\newblock In M.~F. Balcan and K.~Q. Weinberger, editors, {\em Proceedings of
  The 33rd International Conference on Machine Learning}, volume~48 of {\em
  Proceedings of Machine Learning Research}, pages 2915--2923, New York, New
  York, USA, 20--22 Jun 2016. PMLR.

\bibitem{Alexey-19}
S.~Alexey and A.~I. Panov.
\newblock Hierarchical reinforcement learning with clustering abstract
  machines.
\newblock In S.~O. Kuznetsov and A.~I. Panov, editors, {\em Artificial
  Intelligence}, pages 30--43, Cham, 2019. Springer International Publishing.

\bibitem{2007_aspinall}
M.~Aspinall and R.~Hamermesh.
\newblock Realizing the promise of personalized medicine.
\newblock {\em Harvard business review}, 85:108--17, 165, 11 2007.

\bibitem{berner2019dota}
C.~Berner, G.~Brockman, B.~Chan, V.~Cheung, P.~D{\k{e}}biak, C.~Dennison,
  D.~Farhi, Q.~Fischer, S.~Hashme, C.~Hesse, et~al.
\newblock Dota 2 with large scale deep reinforcement learning.
\newblock {\em arXiv preprint arXiv:1912.06680}, 2019.

\bibitem{Bock2007}
H.-H. Bock.
\newblock {\em Clustering Methods: A History of k-Means Algorithms}, pages
  161--172.
\newblock Springer Berlin Heidelberg, Berlin, Heidelberg, 2007.

\bibitem{CORONATO2020101964}
A.~Coronato, M.~Naeem, G.~{De Pietro}, and G.~Paragliola.
\newblock Reinforcement learning for intelligent healthcare applications: A
  survey.
\newblock {\em Artificial Intelligence in Medicine}, 109:101964, 2020.

\bibitem{Dietterich-00}
T.~Dietterich.
\newblock Hierarchical reinforcement learning with the maxq value function
  decomposition.
\newblock {\em The Journal of Artificial Intelligence Research (JAIR)}, 13, 12
  2000.

\bibitem{10.1016/j.jss.2016.02.008}
S.~Ferretti, S.~Mirri, C.~Prandi, and P.~Salomoni.
\newblock Automatic web content personalization through reinforcement learning.
\newblock {\em Journal of Systems and Software}, 121:157--169, 2016.

\bibitem{Ghavamzadeh-01}
M.~Ghavamzadeh and S.~Mahadevan.
\newblock Continuous-time hierarchical reinforcement learning.
\newblock In {\em ICML}, 2001.

\bibitem{10.1016/S0167-7799(01)01814-5}
G.~S. Ginsburg and J.~J. McCarthy.
\newblock Personalized medicine: revolutionizing drug discovery and patient
  care.
\newblock {\em Trends in Biotechnology}, 19(12):491--496, 2001.

\bibitem{Giunchiglia-92}
F.~Giunchiglia and T.~Walsh.
\newblock A theory of abstraction.
\newblock {\em Artificial Intelligence}, 57(2):323--389, 1992.

\bibitem{10.1038/s41591-018-0310-5}
O.~Gottesman, F.~Johansson, M.~Komorowski, A.~Faisal, D.~Sontag,
  F.~Doshi-Velez, and L.~A. Celi.
\newblock {Guidelines for reinforcement learning in healthcare}.
\newblock {\em Nature Medicine}, 25(1):16--18, 2019.

\bibitem{10.5013/IJSSST.a.17.24.06}
X.~Guo and Y.~Zhai.
\newblock K-means clustering based reinforcement learning algorithm for
  automatic control in robots.
\newblock {\em International Journal of Simulation Systems, Science and
  Technology}, 17, 01 2016.

\bibitem{Guo-16}
X.~Guo and Y.~Zhai.
\newblock K-means clustering based reinforcement learning algorithm for
  automatic control in robots.
\newblock {\em International Journal of Simulation Systems, Science and
  Technology}, 17, 01 2016.

\bibitem{HernandezGardiol-00}
N.~Hernandez-Gardiol and S.~Mahadevan.
\newblock Hierarchical memory-based reinforcement learning.
\newblock In {\em NIPS}, 2000.

\bibitem{Kolter-09}
J.~Z. Kolter and A.~Ng.
\newblock Regularization and feature selection in least-squares temporal
  difference learning.
\newblock In {\em ICML '09}, 2009.

\bibitem{Konidaris-09}
G.~Konidaris and A.~Barto.
\newblock Efficient skill learning using abstraction selection.
\newblock In {\em Twenty-First International Joint Conference on Artificial
  Intelligence}, pages 1107--1112, 01 2009.

\bibitem{langford2007epoch}
J.~Langford and T.~Zhang.
\newblock The epoch-greedy algorithm for contextual multi-armed bandits.
\newblock {\em Advances in neural information processing systems}, 20(1):96--1,
  2007.

\bibitem{lasalvia2020personalization}
L.~Lasalvia.
\newblock Personalization and standardization: Can we have it all?
\newblock {\em Journal of Precision Medicine| Volume}, 6(1), 2020.

\bibitem{10.1007/978-3-540-95995-3_14}
H.~Lee and A.~Borodin.
\newblock Cluster based personalized search.
\newblock In {\em International Workshop on Algorithms and Models for the
  Web-Graph}, volume 5427, pages 167--183, 02 2009.

\bibitem{Li-06}
L.~Li, T.~J. Walsh, and M.~L. Littman.
\newblock Towards a unified theory of state abstraction for mdps.
\newblock {\em ISAIM}, 4:5, 2006.

\bibitem{Ma-19}
X.~Ma, S.-Y. Zhao, and W.-J. Li.
\newblock Clustered reinforcement learning, 2019.

\bibitem{Mandel-16}
T.~Mandel, Y.-E. Liu, E.~Brunskill, and Z.~Popovic.
\newblock Efficient bayesian clustering for reinforcement learning.
\newblock In {\em IJCAI}, 2016.

\bibitem{Mannor-04}
S.~Mannor, I.~Menache, A.~Hoze, and U.~Klein.
\newblock Dynamic abstraction in reinforcement learning via clustering.
\newblock {\em Proceedings of the 21st International Conference on Machine
  Learning}, 09 2004.

\bibitem{mnih2016asynchronous}
V.~Mnih, A.~P. Badia, M.~Mirza, A.~Graves, T.~Lillicrap, T.~Harley, D.~Silver,
  and K.~Kavukcuoglu.
\newblock Asynchronous methods for deep reinforcement learning.
\newblock In {\em International conference on machine learning}, pages
  1928--1937. PMLR, 2016.

\bibitem{mnih2013playing}
V.~Mnih, K.~Kavukcuoglu, D.~Silver, A.~Graves, I.~Antonoglou, D.~Wierstra, and
  M.~Riedmiller.
\newblock Playing atari with deep reinforcement learning.
\newblock {\em arXiv preprint arXiv:1312.5602}, 2013.

\bibitem{Pateria-21}
S.~Pateria, B.~Subagdja, A.~Tan, and C.~Quek.
\newblock Hierarchical reinforcement learning: {A} comprehensive survey.
\newblock {\em {ACM} Comput. Surv.}, 54(5):109:1--109:35, 2021.

\bibitem{10.1038/s41539-019-0054-0}
R.~L. Peach, S.~N. Yaliraki, D.~Lefevre, and M.~Barahona.
\newblock {Data-driven unsupervised clustering of online learner behaviour}.
\newblock {\em npj Science of Learning}, 4(1):14, 2019.

\bibitem{Rafati-19}
J.~Rafati and D.~Noelle.
\newblock Learning representations in model-free hierarchical reinforcement
  learning.
\newblock In {\em AAAI}, 2019.

\bibitem{Ricci2011IntroductionTR}
F.~Ricci, L.~Rokach, and B.~Shapira.
\newblock Introduction to recommender systems handbook.
\newblock In {\em Recommender Systems Handbook}, 2011.

\bibitem{Schulman-15}
J.~Schulman, S.~Levine, P.~Moritz, M.~I. Jordan, and P.~Abbeel.
\newblock Trust region policy optimization.
\newblock {\em CoRR}, abs/1502.05477, 2015.

\bibitem{schulman2017proximal}
J.~Schulman, F.~Wolski, P.~Dhariwal, A.~Radford, and O.~Klimov.
\newblock Proximal policy optimization algorithms.
\newblock {\em arXiv preprint arXiv:1707.06347}, 2017.

\bibitem{Strehl-09}
A.~L. Strehl, L.~Li, and M.~L. Littman.
\newblock Reinforcement learning in finite mdps: Pac analysis.
\newblock {\em Journal of Machine Learning Research}, 10(84):2413--2444, 2009.

\bibitem{sutton2018reinforcement}
R.~S. Sutton and A.~G. Barto.
\newblock {\em Reinforcement learning: An introduction}.
\newblock MIT press, 2018.

\bibitem{Tang-16}
H.~Tang, R.~Houthooft, D.~Foote, A.~Stooke, X.~Chen, Y.~Duan, J.~Schulman,
  F.~De~Turck, and P.~Abbeel.
\newblock \#exploration: A study of count-based exploration for deep
  reinforcement learning.
\newblock {\em 31st Conference on Neural Information Processing Systems
  (NIPS)}, 30:1--18, 2017.

\bibitem{Thanh-13}
T.~Thanh, Z.~Li, T.~Silander, and T.-Y. Leong.
\newblock Online feature selection for model-based reinforcement learning.
\newblock {\em 30th International Conference on Machine Learning, ICML}, pages
  498--506, 01 2013.

\bibitem{thomas2015notation}
P.~S. Thomas and B.~Okal.
\newblock A notation for {M}arkov decision processes.
\newblock {\em arXiv preprint arXiv:1512.09075}, 2015.

\bibitem{10.1145/3318236.3318249}
A.~Vatian, S.~Dudorov, A.~Ivchenko, K.~Smirnov, E.~Chikshova, A.~Lobantsev,
  V.~Parfenov, A.~Shalyto, and N.~Gusarova.
\newblock Design patterns for personalization of healthcare process.
\newblock In {\em Proceedings of the 2019 2nd International Conference on
  Geoinformatics and Data Analysis}, ICGDA 2019, page 83–88, New York, NY,
  USA, 2019. Association for Computing Machinery.

\bibitem{vinyals2019grandmaster}
O.~Vinyals, I.~Babuschkin, W.~M. Czarnecki, M.~Mathieu, A.~Dudzik, J.~Chung,
  D.~H. Choi, R.~Powell, T.~Ewalds, P.~Georgiev, et~al.
\newblock Grandmaster level in {StarCraft II} using multi-agent reinforcement
  learning.
\newblock {\em Nature}, 575(7782):350--354, 2019.

\bibitem{10.1145/2623372}
X.~Wang, Y.~Wang, D.~Hsu, and Y.~Wang.
\newblock Exploration in interactive personalized music recommendation: A
  reinforcement learning approach.
\newblock {\em ACM Trans. Multimedia Comput. Commun. Appl.}, 11(1), Sept. 2014.

\bibitem{Wookey-15}
D.~S. Wookey and G.~D. Konidaris.
\newblock Regularized feature selection in reinforcement learning.
\newblock {\em Machine Learning}, 100:655–676, 2015.

\bibitem{10.1109/TNNLS.2020.2975035}
M.~Yang, W.~Huang, W.~Tu, Q.~Qu, Y.~Shen, and K.~Lei.
\newblock Multitask learning and reinforcement learning for personalized dialog
  generation: An empirical study.
\newblock {\em IEEE Transactions on Neural Networks and Learning Systems},
  32(1):49--62, 2021.

\bibitem{pmlr-v85-yauney18a}
G.~Yauney and P.~Shah.
\newblock Reinforcement learning with action-derived rewards for chemotherapy
  and clinical trial dosing regimen selection.
\newblock In F.~Doshi-Velez, J.~Fackler, K.~Jung, D.~Kale, R.~Ranganath,
  B.~Wallace, and J.~Wiens, editors, {\em Proceedings of the 3rd Machine
  Learning for Healthcare Conference}, volume~85 of {\em Proceedings of Machine
  Learning Research}, pages 161--226. PMLR, 17--18 Aug 2018.

\bibitem{Yu2019ReinforcementLI}
C.~Yu, J.~Liu, and S.~Nemati.
\newblock Reinforcement learning in healthcare: A survey.
\newblock {\em ArXiv}, abs/1908.08796, 2019.

\bibitem{10.1145/2566486.2567991}
Y.~Yue, C.~Wang, K.~El-Arini, and C.~Guestrin.
\newblock Personalized collaborative clustering.
\newblock In {\em Proceedings of the 23rd International Conference on World
  Wide Web}, WWW '14, page 75–84, New York, NY, USA, 2014. Association for
  Computing Machinery.

\bibitem{10.1007/s11257-021-09292-w}
Y.~Zhang and W.-B. Goh.
\newblock Personalized task difficulty adaptation based on reinforcement
  learning.
\newblock {\em User Modeling and User-Adapted Interaction}, pages 1--32, 2021.

\bibitem{Zhao2009ReinforcementLD}
Y.~Zhao, M.~Kosorok, and D.~Zeng.
\newblock Reinforcement learning design for cancer clinical trials.
\newblock {\em Statistics in medicine}, 28 26:3294--315, 2009.

\end{thebibliography}
%%%%%%%%%%%%%%%%%%%%%%%%%%%%%%%%%%%%%%%%%%%%%%%%%%%%%%%%%%%%%%%%%%%%%%%%%%%%%%%%%%%%%%%%%%%%%%%%%%%%%%%

\end{document}